# A Unified Python Framework for Direct PPO-based Control of AHUs with Economizer Logic and $CO_2$-Constrained Ventilation

Erfan Haghighat Damavandi[a], Davide Papurello[b], Mahdi Alibeigi[c], Armin Keshavarz[d], Simone Canevarolo[b], Marco Condo[b]

[a] *Research and Development Department Owtana Tech*- Turin, Italy
[b] *Department of Energy Politecnico di Torino*- Turin, Italy
[c] *Department of Energy University of Tehran*- Tehran, Iran
[d] *Department of Energy Sharif University of Technology*- Tehran, Iran

---

*Abstract*: Optimizing HVAC (Heating, Ventilation and Air Conditioning) can enhance a building's energy efficiency while providing comfort levels for its occupants. Using conventional control systems to maintain HVAC functions is often difficult because of the nonlinear characteristics of a building envelope as it experiences stochastic load variations over time. This paper presents a new approach to optimizing HVAC systems through the use of Deep Reinforcement Learning (DRL) algorithms and the Proximal Policy Optimization (PPO) algorithm implemented in a custom Python performance environment. The DRL system uses a second-order resistor-capacitor thermal model and an integrated dynamic mass balance of $CO_2$ to replicate the complex physics associated with buildings. One major innovation of this study is a "Hierarchical Flow Logic," which provides the means to ensure that indoor air quality (IAQ) is maintained by overriding the accepted actions of the agent that cause $CO_2$ to exceed 1000 ppm. In addition, an enthalpy-based economiser is used to create free cooling from the outdoor environment. The experimental data shows that compared to PID controllers tuned by GA or traditional On-Off controls, a PPO agent has better temperature stability and energy efficiency overall. An end-to-end pipeline provides an avenue for robust and generalized solutions to help implement smart building energy management within the context of real hardware implementations.

*Keywords: Reinforcement Learning, PPO Algorithm, HVAC Control, Indoor Air Quality (IAQ), 2R-2C Thermal Model, Energy Efficiency, Python-based Simulation, Occupancy-driven Ventilation, Multi-objective Optimization.*

## I. INTRODUCTION

The majority of energy used worldwide comes from buildings, and HVAC systems (potentially the largest contributor to energy consumption). Improving how efficient HVAC systems operate has the greatest potential for reducing energy use and waste in commercial buildings. While current metrics used to measure the efficiency of equipment like HVAC systems (for example, the EER (energy efficiency ratio) and COP (coefficient of performance) are well-established, they do not always represent how well HVAC systems work in the real world, which has a large degree of variability and is highly complex.

Buildings account for approximately 40 % of total global energy use, with HVAC systems making up between 30 to 40 percent of that energy consumption [1], [2], [3].

The Air Handling Unit (AHU) is an essential element of the operation that provides control over temperature, humidity and the supply of fresh air to the building. As a result of this fact, the AHU has a direct effect on both the thermal comfort level of building occupants and the total energy consumed. Humidity control is one critical factor in the overall environmental control; in addition to influencing perceived air quality and Sick Building Syndrome [4], it can also represent a sizable thermodynamic load. Failure to properly control the amount of water vapour in a controlled environment can result in negative health outcomes, as well as high energy costs due to cooling coil heat load.

The development of building simulation technology and occupant monitoring has produced more accurate training environments than previously available because emphasis has been placed on reducing total energy usage within building space. Reducing total energy usage in buildings but still meeting the comfort levels of occupants, particularly with respect to their thermal comfort and indoor air quality (IAQ) is pressing concern for many [5], [6] and may be difficult (non-linear) when considering the relationship that exists among temperature, humidity, $CO_2$ concentration and dynamic occupancy patterns via traditional control strategies such as employing on/off thermostats and using basic control strategies (PID controllers) using fixed set points or rule-based heuristics to control targets. Consequently, this results in significant amounts of wasted energy and/or discomfort to occupants.) In recent years, reinforcement learning (RL), has emerged to be viable alternative approach to data driven HVAC control as RL learns how to optimize a policy through direct interaction with the real-world environment using trial & error rather than requiring an explicit physical model as is required by Model based Predictive Control [1]. Among reinforcement learning algorithms, Proximal Policy Optimization (PPO) has developed considerable recognition because it is stable and efficient when dealing with continuous control tasks. One example is the successful implementation of PPO by Islam et al. [7] to schedule electric vehicle charging and water heating to save approximately 30% in costs in a residence. Additionally, Chatterjee and Khovalyg [8] reviewed deep reinforcement learning (DRL) techniques (including PPO) to design dynamic indoor thermal environments that produce a balance between energy consumption, thermal comfort and potentially long-term occupant health, assuming that an appropriate reward function is established. Current research by Chatterjee and Khovalyg [9] has illustrated an example of a DRL controller (DIET), which varies indoor temperature dynamically and reduced heating energy consumption by 28%-64%, while providing a dynamic environment for 96% of occupied hours. Haghighat and Zhang [10] discussed the use of deep learning along with digital twins to predict indoor environmental conditions, while Yun et al. [11] provided a comprehensive multi domain review on non-intrusive, AI based occupant monitoring for thermal, air and visual comfort. Mohsenpour and Xing [12] introduced a co adaptive RL approach that incorporates occupant feedback directly into the framework and combines RL with model predictive control (MPC) to co optimize energy use, comfort and indoor air quality (IAQ). Hashir et al. [13] conducted a systematic review of AI based IAQ monitoring in educational institutions, identifying the transition away from conventional, reactive approaches to proactively controlling the building's ventilation using AI as the basis for doing so. Additionally, with regards to hybrid ventilation systems. According to Ahrendsen et al. [14], Monte Carlo Simulation with sensitivity analysis can be used to characterize controls space and determine a few key control parameters. Li et al. [15] demonstrated how smart thermostats' occupancy patterns and setpoint preferences were highly variable. This suggests an urgent need for data driven control strategies that can be dynamically adjusted.

These advancements have left major gaps. Most RL studies in HVAC (i) treat the quality of indoor air (IAQ) as a penalty, rather than a primary objective, (ii) are dependent on co-simulation tools that add unnecessary complication to how easy they can be implemented and, (iii) use reward functions that do not enforce IAQ constraints in a hard way. Further, it is uncommon to find integrated python-based environments with physical models such as 2R 2C thermal network and dynamic balance mass of $CO_2$ built-in with a policy optimization (PPO) agent. The main contribution of this research is to apply a direct reinforcement learning approach to the HVAC system.

This paper presents a unified Python framework for direct PPO based control of an AHU to address research gap. The main contributions are:

- A self-contained Gym style environment that couples a 2R 2C thermal model and a dynamic $CO_2$ balance, eliminating the need for external $CO_2$-simulation.
- The sole goal of the RL agent is temperature regulation, with the temperature deviation from the 22 °C setpoint being evaluated using squared errors as penalties.
- Mass balance calculation is available for determining the minimum volume of fresh air required for $CO_2$ dilution.

This paper is organized as follows. Section 2 describes the physical models, the 2R-2C thermal network, moisture balance, and $CO_2$ dynamics. Section 3 details the baseline thermostatic (On/Off) control strategy, while Section 4 presents the advanced Proportional-Integral-Derivative (PID) controller. Section 5 introduces the proposed PPO based RL framework, including state/action spaces, reward engineering, and hierarchical ventilation logic. Section 6 reports simulation results and compares the performance of all controllers.

## II. Physical and Mathematical System Modelling

This research evaluates three dynamic models to simulate comfort: a Thermal Network, a Moisture Balance and an Indoor Air Quality Model; it incorporates a 2R-2C dynamic model (see Fig.1) that uses empirical thermodynamic constants as well as a 2R-1C representation of the building's envelope to enable accurate real-time representations of transient thermodynamic and hygrometric behavior. By creating this framework, the performance of the HVAC system can be analyzed for 24 hours, taking the effects of ventilation, moisture transfer and contaminant levels into account.

**Table. 1.** Input Parameters and Environmental Variables of air Properties and Building Geometry

| Parameter | Definition | Value |
|---|---|---|
| $c_p$ | Specific heat capacity of air | 1005 J/(kg·K) |
| $\rho_{air}$ | Air density | 1.2 kg/m$^3$ |
| $V_{house}$ | Building volume | 1200 m$^3$ |
| $M_{air,eff}$ | Effective air mass (includes furnishings) | $V \times \rho \times 10$ |
| $h_{fg}$ | Latent heat of vaporization of water | 2,501,000 J/kg |
| $C_{wall}$ | Wall thermal capacitance | $10^7$ J/K |
| $R_{out}$ | External thermal resistance | 0.02 K/W |
| $R_{in}$ | Internal thermal resistance | 0.05 K/W |
| $RH_{out}$ | Outdoor Relative Humidity | 30 to 60 % |
| $T_{supply}$ | Supply air temperature from AHU | 12°C |
| $T_{out}$ | Outdoor Temperature | 20°C-30°C |
| $W_{supply}$ | Supply air humidity ratio | 0.007 kg_water/kg_dry air |
| $\dot{V}_{max}$ | Maximum supply airflow rate | 8 m$^3$/s |
| $A$ | Fresh air ratio (fixed in baseline) | 0.5 (50%) |
| $CO_{2,out}$ | Outdoor $CO_2$ concentration | 400 ppm |
| $CO_{2,limit}$ | Acceptable indoor $CO_2$ limit | 1000 ppm |
| $G_{CO_2}$ | $CO_2$ generation rate per person | 0.01 L/(s·person) |
| $\dot{m}_{moisture}$ | Moisture generation rate per person | 2×10$^{-5}$ kg/(s·person) |
| $\dot{Q}_{in}$ | Internal Heat Gain | 25,000 to 70,000 W |

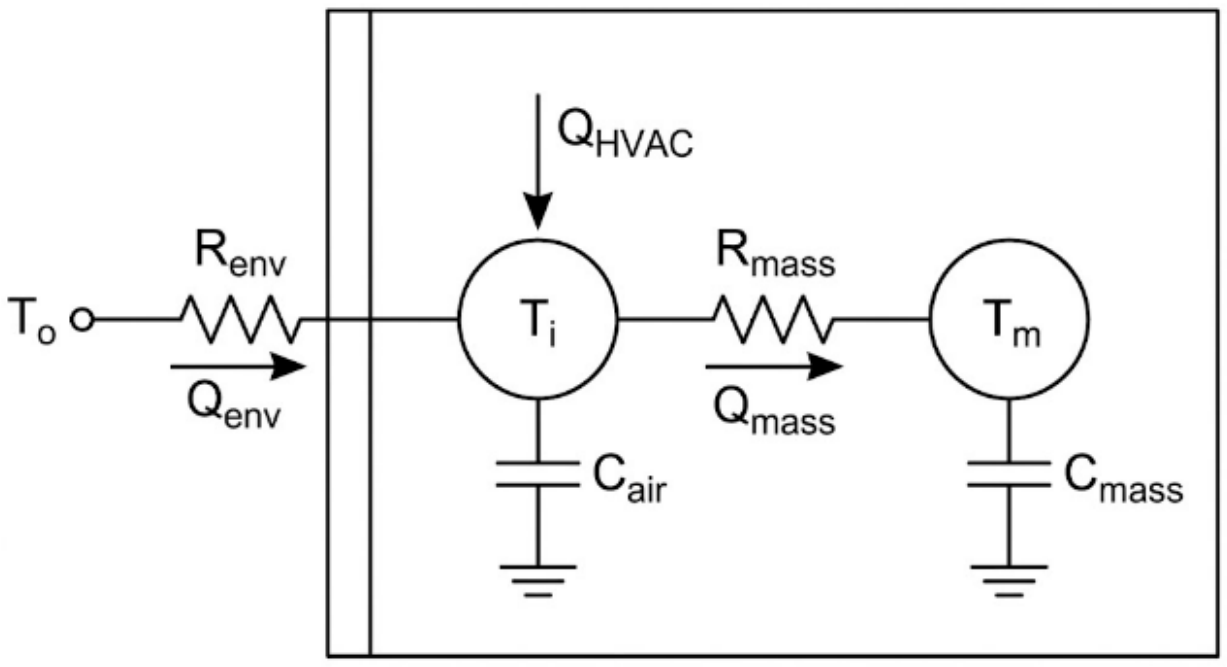

**Fig. 1.** 2R-2C model of space

The wall temperature is updated using thermal resistance relationships.

$$\frac{dT_{wall}}{dt} = \frac{\frac{T_{out} - T_{wall}}{R_{out}} + \frac{T_{air} - T_{wall}}{R_{in}}}{C_{wall}} \quad (1)$$

where $T_{\text{out}}$ is outdoor air temperature and $T_{\text{air}}$ is indoor air temperature.

The indoor air temperature is updated using an energy balance.

$$\frac{dT_{air}}{dt} = \frac{\frac{T_{wall} - T_{air}}{R_{in}} + Q_{internal} - Q_{cooling}}{M_{air}c_p} \quad (2)$$

$Q_{internal}$ represents internal heat gains (people, equipment, lighting). $Q_{\text{cooling}}$ is the sensible cooling delivered to the zone.

Indoor humidity is updated using a mass balance equation according to:

$$\frac{dw}{dt} = \frac{\dot{m}_{moisture} + \rho_{air}.\dot{V}\left(W_{supply} - W_{air}\right)}{M_{air}} \quad (3)$$

where $\dot{m}_{\text{moisture}}$, $\dot{V}_{\text{total}}$, and $W_{\text{supply}}$ are moisture generation rate by occupants (kg/s), total volumetric airflow rate (m³/s) – the control action, and humidity ratio of supply air (fixed = 0.007 kg/kg), respectively.

Indoor $CO_2$ concentration is calculated using a dynamic balance.

$$\frac{dCO_2}{dt} = \frac{G_{CO_2} + \dot{V}_{fresh}\left(CO_{2,out} - CO_{2,room}\right)}{V} \quad (4)$$

where $C_{CO_2}$ , $G_{CO_2}$ , $\dot{V}_{\text{fresh}}$ , and $C_{\text{out}}$ indoor $CO_2$ concentration (ppm), $CO_2$ generation rate per person = (0.01 L/s) [based on ASHRAE 62.1], fresh air flow rate $\dot{V}_{fresh} = \alpha \dot{V}_{\text{total}}$ with $\alpha = 0.4$ (baseline), outdoor $CO_2$ concentration = 400 ppm, respectively.

Supply air temperature is fixed at $T_{\text{supply}} = 12\,^{\circ}C$. The mixed air humidity ratio before the cooling coil is:

$$W_{mixed} = (1 - \alpha)W_{return} + \alpha W_{outdoor} \quad (5)$$

where $W_{return}$ is the return air humidity ratio (taken equal to the zone humidity ratio $w$), and $W_{\text{outdoor}}$ is the outdoor humidity ratio derived from outdoor $T$ and $RH$.

The total cooling load ($Q_{coil}$) consists of sensible cooling and latent cooling that calculated as:

$$Q_{sensible} = \dot{V}\rho c_p(T_{mixed} - T_s) \quad (6)$$

$$Q_{latent} = \dot{V}\rho h_{fg}\left(W_{mixed} - W_{supply}\right) \quad (7)$$

$$Q_{coil} = Q_{sensible} + Q_{latent} \quad (8)$$

To test the controller under realistic daily variations, the following profiles are used (period = 24 h, $t$ in seconds):

**Table. 2.** Input Parameters and Environmental Variables of air Properties and Building Geometry

| Variable | Equation | Range | Unit |
|---|---|---|---|
| $T_{out}$ | $T_{out} = 25 + 5\,sin(2\pi t/86400)$ | 20–30 | °C |
| $RH_{out}$ | $RH_{out} = 0.3 + 0.3\,sin(2\pi t/86400)$ | 30–60 | % |
| Occupancy | $N = 70 + 80\,sin(2\pi t/86400)$ | 70–150 | persons |
| $\dot{Q}_{in}$ | $Q_{\text{internal}} = 25000 + 45000\,sin(2\pi t/86400)$ | 25000–70000 | W |

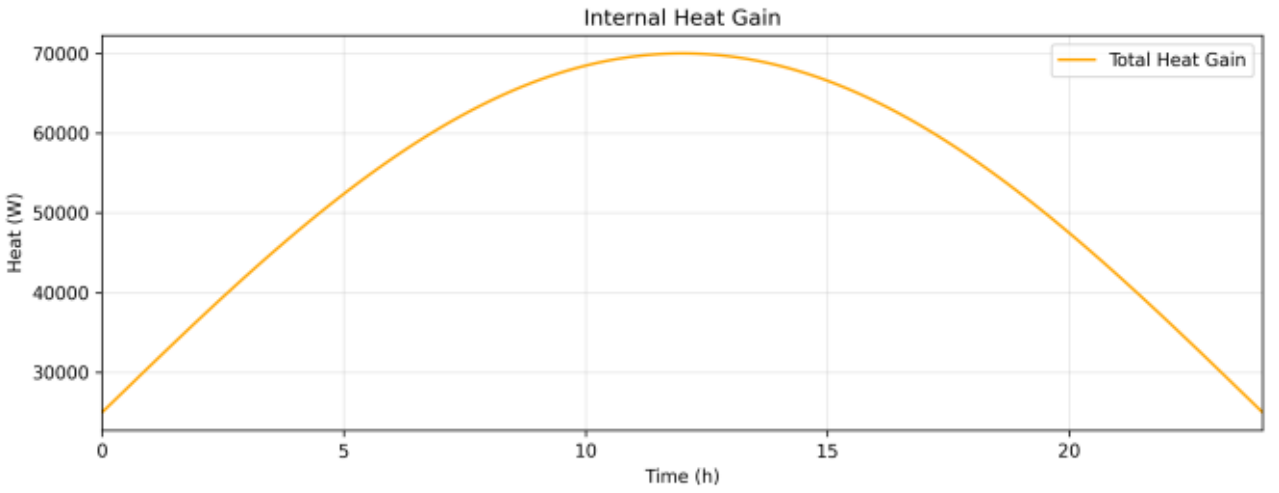

**Fig. 2.** Internal heat gain over the time

Total energy consumption is obtained by integrating coil power over time:

$$E_{total}[KWh] = \frac{1}{3.6 \times 10^6} \int_{t_0}^{t_f} Q_{Coil}(t).\,dt \quad (9)$$

Occupants generate moisture through respiration and perspiration. To simplify psychrometric calculations, the saturation vapor pressure is calculated using Buck's empirical relation:

$$P_{sat} = 611.21 \cdot exp\left(\frac{18.678 - \frac{T}{234.5}}{T + 257.14}.T\right) \quad (10)$$

Where Wall thermal capacitance

$$w = 0.622\,\frac{(RH/100) \cdot P_{sat}(T)}{P_{atm} - (RH/100) \cdot P_{sat}(T)} \quad (11)$$

Then, the humidity ratio $w$ (kg water / kg dry air) is calculated from relative humidity (RH in %) and atmospheric pressure $P_{\text{atm}} = 101325$ Pa (at sea level).

## III. Baseline Control Strategies

*3.1 Conventional Thermostatic (On-Off) Control*

A regular thermostatic control system is used to provide a comparison some ways of measuring performance. An on/off thermostat represents how to totally control residential digital characters with temperature ranges that are defined by a continuous setting (or point) and a fixed differential value (or hysteresis).

In operation, AHUs will work at full load until the lower bound of temperature, where AHU will stop until the temperature reaches an upper threshold. This approach may be simple, but it can create large temperature oscillations and also, will not give a granular level of control on the fresh air ratio to provide adequate Indoor Air Quality (IAQ).

**Control Logic:**

The controller works to maintain a temperature setpoint of $T_{set}$ with a range of tolerance equal to the differential. The mode of fan operation $(S_{fan})$ will be determined by the following switching logic:

$$S_{fan}(t) = \begin{cases} 1 & \text{if } T_{air} > T_{set} + \dfrac{\Delta T}{2} \\ 0 & \text{if } T_{air} < T_{set} - \dfrac{\Delta T}{2} \\ S_{fan}(t-1) & \text{otherwise} \end{cases} \qquad (12)$$

In this simulation, the setpoint was established at 22°C with a 2°C differential, meaning the cooling system activates when the indoor temperature exceeds 23°C and deactivates when it drops below 21°C.

## IV. Advanced Baseline: Proportional-Integral-Derivative (PID) Control

An additional evaluation of the proposed reinforcement learning (RL) agent's performance compared to existing industry standards required implementing a proportional-integral-derivative (PID) controller. A thermostat uses a discrete switching mechanism to provide ON/OFF control. The PID uses continuous control signals, making it capable of providing smoother modulation of fan speeds and achieving more accurate temperature control [11].

4.1 Control Strategy and Formula A PID controller calculates the actuation signal (u(t)) to the fan by continuously measuring the error (e(t)).

$$e(t) = T_{air} - T_{set} \qquad (13)$$

The control output is the sum of three distinct terms [16], [17]:

$$u(t) = K_p e(t) + K_i \int e(t)dt + K_d \frac{de(t)}{dt} \qquad (14)$$

where $K_p$ , $K_i$ $K_d$ are the proportional, integral, and derivative gains with amount of 0.5, 0.001, 0.1, respectively.

*4.2 Actuator Saturation and Modulation*

The raw output of the PID formula u(t) is mapped to the physical limits of the fan. The fan cannot run in a reverse direction. The fan also cannot exceed maximum speed. Clipping (saturation) function applies here too; hence we have two cases:

$$S_{fan} = min\left(1, max\left(0, u(t)\right)\right) \qquad (15)$$

The final output of the system is a signal that indicates the fan ratio, which is scaled proportionally between 0 and 1 to scale the mass flow rate $\dot{V}_{air}$ to a maximum of $8.0$ $m^3$/s, constantly modulating the fan to balance the 2R-2C physics equations better than a binary switch could do, especially in the high load conditions simulated by ($Q_{internal} \leq 70$ kW). The continuous modulation provides improved balancing of the physics equations of 2R-2C from traditional binary switches, especially during high load conditions.

*4.3 Computational Implementation*

The controller was developed using the programming language Python with a discrete approximation of time. The accumulated error was used to update the error integral once every second, and the derivative was the change in error over two consecutive time steps. This will be the best performing implementation of traditional control systems prior to possible implementations involving Reinforcement Learning [18].

## V. Proposed Reinforcement Learning Framework

Evolution of the HVAC Control Strategy

**Phase 1:** The Baseline Model (Fixed 50% Fresh Air)

In the initial development stage, the system was tested using a standard Reinforcement Learning (PPO) approach with a static operational parameter. The Fresh Air Ratio was locked at a constant 50%.

The PPO agent learned how to vary the total airflow to keep the desired temperature, however, it did not understand its environment. It was forced to supply 50% outside air even if that air was very hot and/or humid, and even when the building was vacant, causing excessive energy use and "thermal stress" on the cooling coil as it frequently had to cool much of the outside air supplied to it.

The second phase of the Smart Systems Project (Economizer and $CO_2$ Balance Control) introduced into an advanced, multi-variable control logic to enhance baseline capabilities. Phase 2 efforts consist of two primary components:

Demand-Controlled Ventilation (DCV) uses real time data of $CO_2$ mass balance to determine the amount of fresh air to bring in, rather than a preset percentage of outside air. By monitoring the level of occupancy inside, the system can determine the amount of fresh air needed to keep indoor pollutants under the safety level of 1000 ppm. This not only creates a high level of IAQ, but also has the effect of minimizing "excess" outside air when occupancy is low.

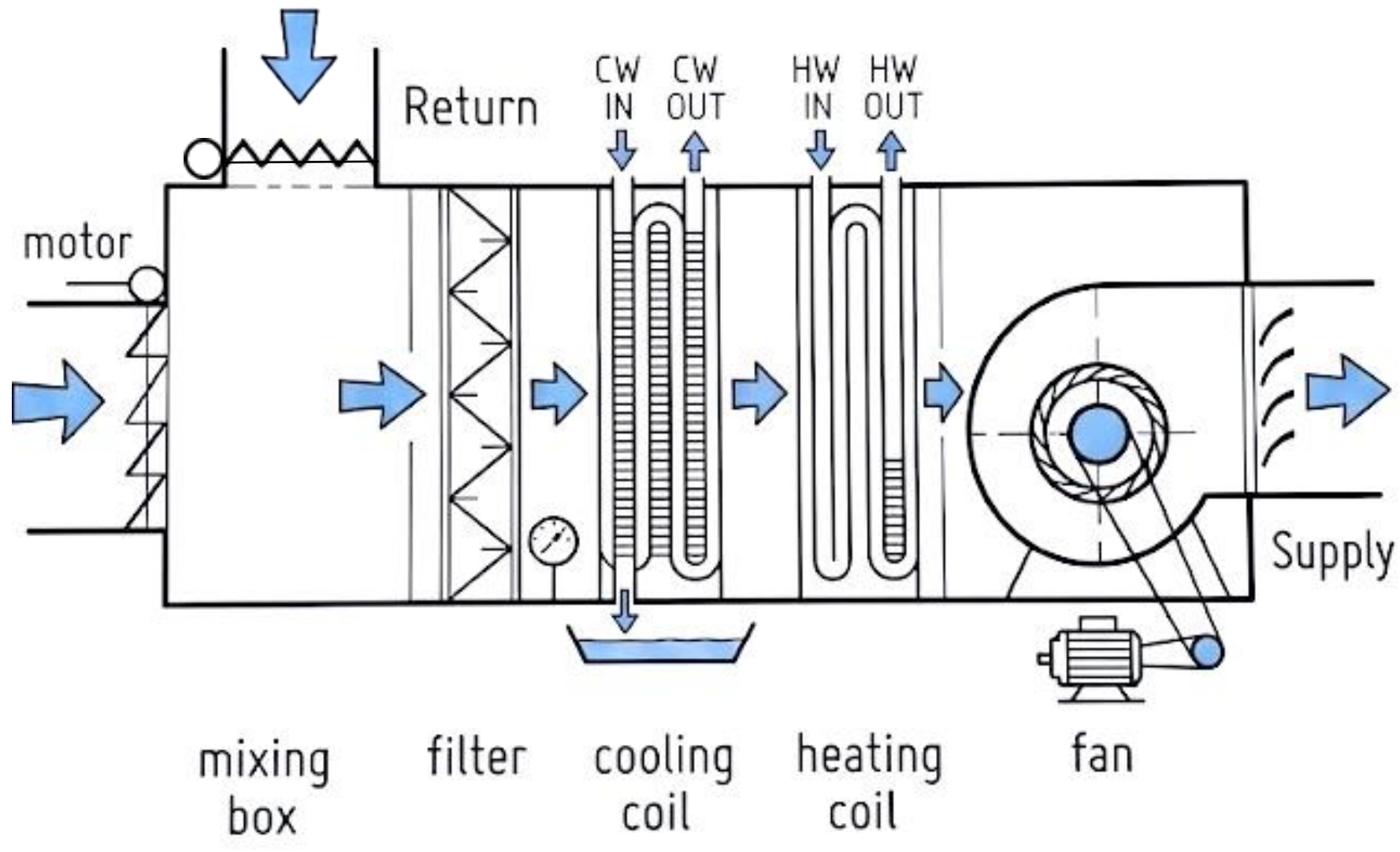

**Fig. 3.** AHU system with fresh air

• Enthalpy-Based Economizer (Free Cooling): The system now compares the Enthalpy (total heat content) of the indoor air versus the outdoor air.

• In economizer mode, the system will override its standard minimum value and can use greater levels of outside air when the outdoor temperature and/or humidity is less than the indoor conditions. This results in lower use of the mechanical cooling system as the environment is used for free cooling.

• Indoor air now has access to outdoor air only when outside conditions (temperature and/or humidity) are colder or drier than inside, in this case, system has limited intake to only whatever is needed for occupants to breathe, essentially sealing the structure's thermal envelope to conserve energy.

Our suggested solution consists of an adapted gym style environment that combines the 2R-2C thermal physics with the Deep Reinforcement Learning agent. The implementation of Stable Baselines 3 is utilized to apply the Proximal Policy Optimization (PPO) algorithm through a single Python model.

5.1 Environment Design and State

The HVACEnv class establishes the building physics environment between an agent and the physical characteristics of a building. Each time step of one second, the agent receives St, an observation from this environment.

According to the _get_obs method in our implementation, the state space is defined as:

$$s_t = [T_{air}, \text{e(t)}] \tag{16}$$

$$e(t) = (T_{set} - T_{air}) \tag{17}$$

where $T_{air}$ is the current indoor temperature and $e(t)$ represents the instantaneous thermal error. The setpoint is fixed at $T_{set} = 22°C$. The compact state space leads to the agent only focusing on keeping the setpoint while observing how far away it is from the target.

5.2 Action Space and Actuation Logic (A)

The agent outputs a continuous action:

$$s_t = [T_{air}, \text{e(t)}] \tag{18}$$

$$a_t \in [-1,1] \tag{19}$$

which is clipped to the physically valid range:

$$u(t) = \text{clip}(a_t, 0,1) = min\,(1, max(0, a_t)) \tag{20}$$

The action produced by this agent will be clipped prior to being converted to a physical mass flow rate using the step function. The Hierarchical Flow Logic is another original design element of our code and functions to balance an agent's decisions against ensuring safety and meeting IAQ requirements.

Agent Flow:

$$\dot{V}_{agent\ flow} = u_t \cdot \dot{V}_{max} \tag{21}$$

IAQ Minimum Flow:

At each timestep, the environment calculates the minimum fresh airflow required to dilute the $CO_2$ generated by the occupants, based on the current people profile (number of occupants over time). Using a steady-state mass balance:

$$\dot{V}_{fresh\ air} = \frac{G_{CO_2}.N_{oc}}{C_{limit} - C_{out}} \tag{22}$$

where $N_{oc}$ is taken directly from the people profile.

Final Actuation:

The actual mass flow ($m_{final}$) is the maximum of the agent's request and the fresh air requirement:

$$m_{final} = max\left(\dot{V}_{agent\ flow},\ \dot{V}_{fresh\ air}\right) \tag{23}$$

Because of this logic, an RL agent is unable to shut off a fan to conserve energy, as the environment will counteract any fan-off attempts made by the RL agent with the desire to remain under 1000 ppm of $CO_2$.

### 5.3 Reward Engineering

Agent reward functions are the primary method of providing reward learning for the RL agent. Our implementation of the RL agent is done with MSE of temperature differential between the sensor and the temperature set-point of the variable (sensor temperature) as the reward function.

$$R = -(|\ T_{air} - T_{set}\ |^2) \tag{24}$$

Therefore, by reducing the negative reward resultant from the coefficient, the PPO agent can learn how to properly change the fan speeds to balance out dynamic heat loads ( $Q_{internal}$ ) and fluctuating outdoor temperatures. The HVACEnv simulates two dynamically interacting physical systems simultaneously.

**Listing. 1.** Python implementation of the custom HVAC environment and PPO training loop

```
import gymnasium as gym
import numpy as np
from stable_baselines3 import PPO

class HVACEnv(gym.Env):
    def __init__(self, ts=22.0):
        self.ts = ts
        B = gym.spaces.Box
        F = np.float32
        self.action_space = B(0, 1, (1,), F)
        self.observation_space = B(-100, 1000, (2,), F)

    def reset(self, seed=None, options=None):
        super().reset(seed=seed)
        self.t = 25.0
        return np.array([self.t, 3.0], np.float32), {}

    def step(self, action):
        self.t += 0.01 * (self.ts - self.t)
        err = self.t - self.ts
        o = np.array([self.t, err], np.float32)
        return o, -(err**2), False, False, {}

env = HVACEnv()
model = PPO("MlpPolicy", env)
model.learn(200000)

obs, _ = env.reset(seed=42)
action, _ = model.predict(
    obs, deterministic=True
)
```

To accomplish this in the code base, there is an important "Supply Temperature Adjustment" technical feature. The fresh air intake $\dot{V}_{fresh\ air}$ is determined based on the agent's flow rate, and if the fresh air intake exceeds the agent's flow rate, the supply temperature $\left(T_{s,a}\right)$ will be dynamically adjusted to keep the two thermal systems at thermal equilibrium:

$$T_{s,a} = T_{air} - \left(\frac{\dot{m}_{agent} \cdot \left(T_{air} - T_{s,n}\right)}{\dot{m}_{total}}\right) \tag{25}$$

This ensures that the physics of the simulation remain consistent even when ventilation overrides the agent's thermal cooling command.

### 5.5 Training Configuration

Training the agent using the Proximal Policy Optimization (PPO) reinforcement learning algorithm with a Multi-Layer Perceptron (MLP) policy occurred over a training period of 300,000 timesteps during which the agent's environment was affected by dynamic disturbances on an hourly basis (24-hour working day). This thorough training protocol will yield a highly generalizable model that is robust enough to handle all building load scenarios.

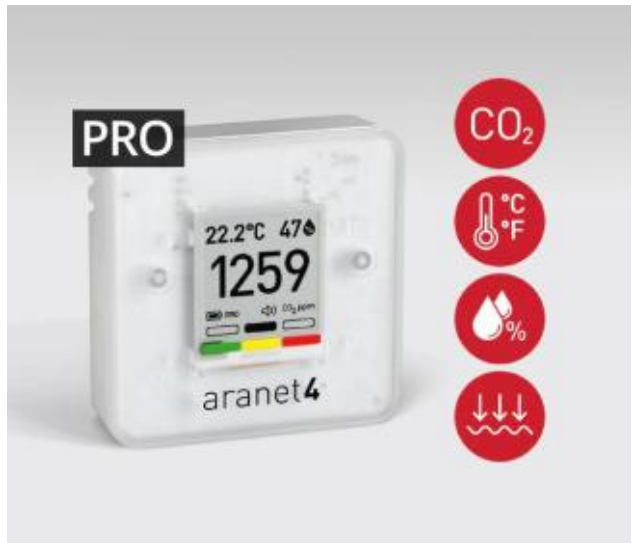

**a.** Aranet4 PRO

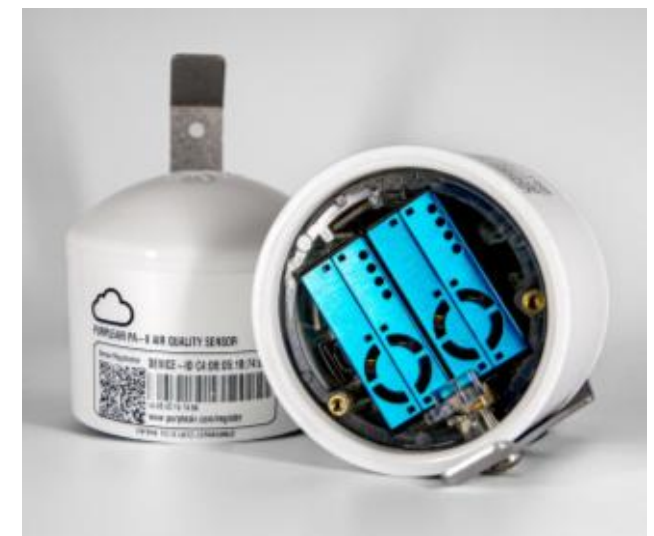

**b**. PurpleAir classic plus

**c.** Air flow meter

**Fig. 3.** Experimental devices used for environmental monitoring: (a) Aranet4 PRO for $CO_2$ and temperature tracking, (b) PurpleAir Classic Plus for air quality assessment, and (c) DAFM4 air flow meter.

### 5.6 System Devices

An environmental monitoring framework includes both indoor/outdoor ambient sensors as well as individual HVAC duct instruments. The Aranet PRO (Fig.3a) is an air-quality monitor for the main hall that includes $CO_2$, temperature and humidity Readins. Meanwhile, the outdoor environmental conditions and air-quality measurements are obtained using the weather-proof PurpleAir Classic Plus sensor station (Fig.3b) that measures air quality. A digital anemometer (air flow meter) (Fig.3c) is installed inside an HVAC Air Supply Duct for measuring airflow and determining the volumetric flow (air moving into or out of) through the fresh and return air ducts, providing a precise calculation of the building's total air exchange efficiency.

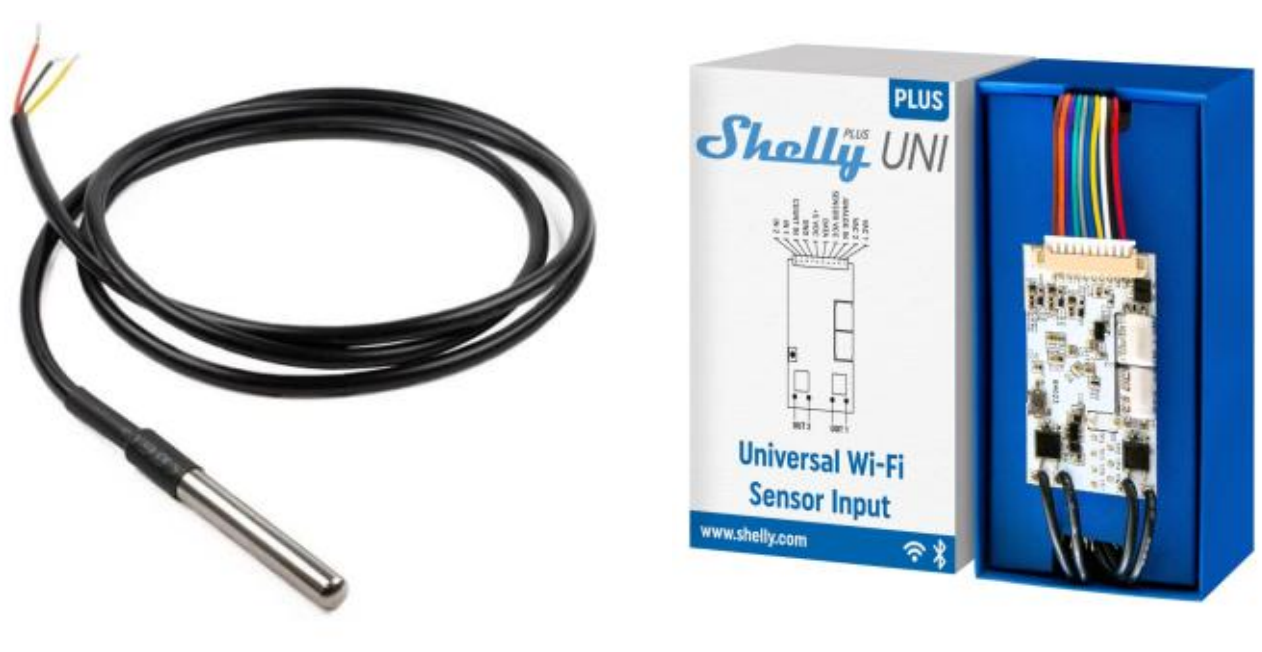

**a.** Temperature sensor **b.** Shelly plus uni

**Fig. 4.** Experimental devices used for temperature monitoring: (a) temperature sensor is based on the DS18B20 chip, (b) Shelly plus uni with universal Wi-Fi sensor input.

The creation of a high-performance data acquisition system which includes a Shelly Plus Uni universal Wi-Fi sensor interface (Fig.4b) to provide an accurate measurement of the thermal properties of an environment. The combination of waterproof and stainless steel temperature probes (Fig.4a), along with a compact microcontroller allows continuous logging of temperature variations throughout an enclosed area. This measurement system was specifically designed to measure the temperature of the internal air in duct work and record both the temperature of the supply water and the return water in the heating and cooling coils while supplying data in real time using Wi-Fi without the need for an independent controller.

In order to measure the thermal capacity of this system, the power to the heating/cooling coil is calculated using the water flow rate entering the system and the temperature difference between the water lines that supply and return from the coil. Simultaneously, the temperature of the incoming and returning air is being monitored by temp sensors mounted in both of the air streams. To characterize the aerodynamics of the air streams, the airflow rate through each of the streams is going to be determined by performing a series of calibration steps. In each of the steps, both the fresh and returned damper blades will be moved a small amount (incrementally) and the resulting flow rate will be recorded for use in an ongoing data program. The collected data will be calculated into a dynamic airflow/damper correlation table that will then be incorporated as part of the project operational logic.

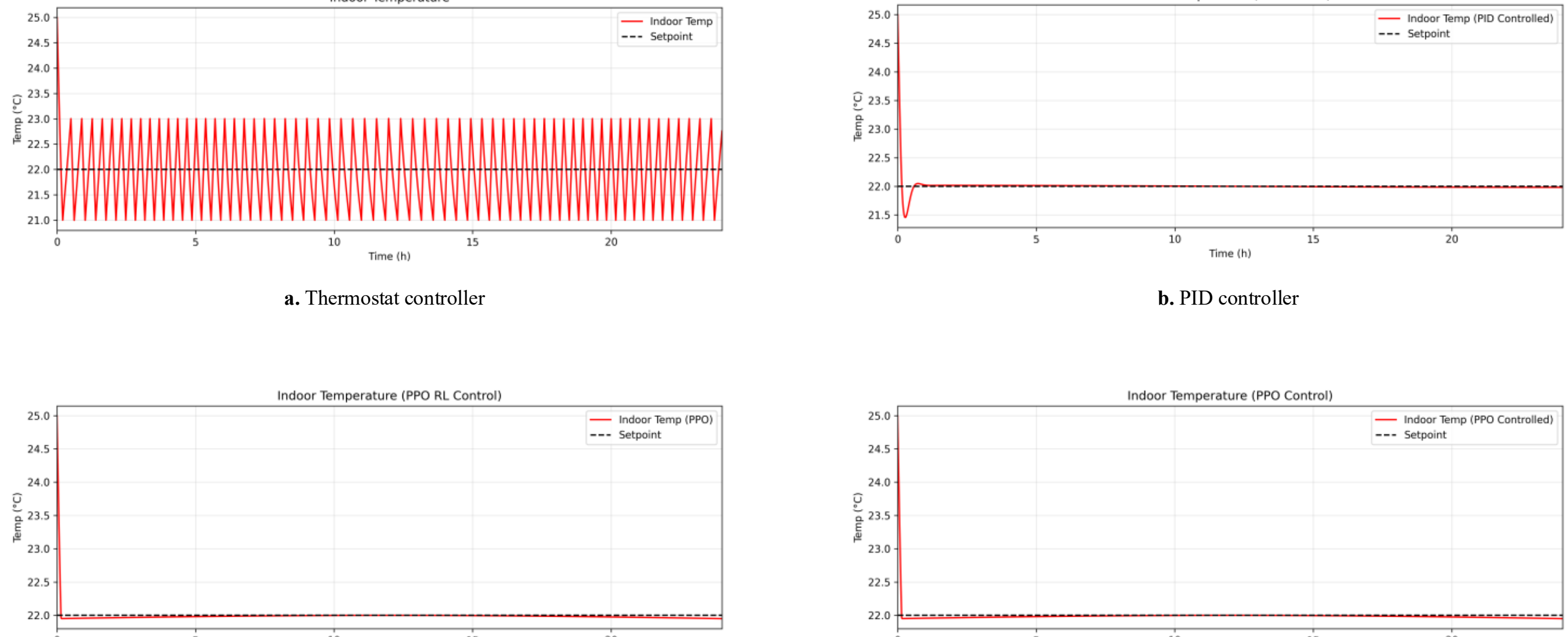

**a.** Thermostat controller **b.** PID controller

**c.** PPO controller **d.** PPO controller with economizer

**Fig. 5.** Comparison of temperature variations under different controllers for an Air Handling Unit system

## VI. RESULTS AND DISCUSSION

The goal of this research was to judge and contrast the efficacy of four separate HVAC monitoring tactics over a 24-hour period in a dynamically loaded building (thermal, moisture, and pollutant $CO_2$ to evaluate how well each can keep the building at comfortable temperatures and indoor air quality (IAQ), while optimizing energy efficiency. Scenarios included: conventional moving the thermostat from "on" or "off" rather than setting to a desired temperature, proportional-integral-derivative (PID) thermostat operation, and PPO Reinforcement learning controller using only positive reinforcement and PPO Reinforcement learning controller with smart fresh air.

6.1. Thermal Comfort Control: (Temperature) On/Off Thermostat A traditional thermostat has a characteristic sawtooth pattern when controlling the air temperature. As seen in (Fig.5a), this behaviour causes a rapid, continuous oscillation around the desired 22-degree Celsius setpoint. The inherent nature of on/off controls leads to significant thermal discomfort for occupants due to large observable fluctuations. Shown in (Fig.5b), the PID controller quickly stabilizes the temperature to the desired setpoint with no oscillation during operation, creating an overall improved level of thermal comfort than with the use of a thermostat.

Two controllers that utilize reinforcement learning (RL) (Fig.5c) and (Fig.5d) methodology also showed fast stabilization of the indoor temperature, produced the same level of stability as the PID controller and outperformed the on/off thermostat by a considerable margin.

6.2. Indoor Air Quality Management: Indoor $CO_2$ dynamics for four control methods relative to a safety threshold of 1000 ppm and a peak occupancy profile are shown in (Fig.6). As seen in (Fig.6a), the thermostat controller creates large fluctuations in indoor $CO_2$ levels, with an increasing number of occurrences of $CO_2$ exceeding 1300 ppm, because of the thermostat's "on-off" control logic. The PID controller (Fig.6b), on the other hand, does not produce any fluctuations due to using modulated ventilation to control indoor $CO_2$ levels, and consequently maintains indoor $CO_2$ levels in the 920 ppm range; however, there are still some instances of this controller exceeding the limit on occupancy drops. As shown in (Fig.6c), the PPO Reinforcement Learning agent provides a more active approach to regulate $CO_2$ levels to remain below 910 ppm at all times during peak demand hours, supporting better indoor air quality. The PPO controller that includes economizer logic (Fig.6d) demonstrates a high degree of accuracy by maintaining precise tracking along the 1000 ppm threshold, thus preventing over ventilation while also providing the dual benefits of ensuring regulatory compliance with indoor air quality and optimized energy efficiency.

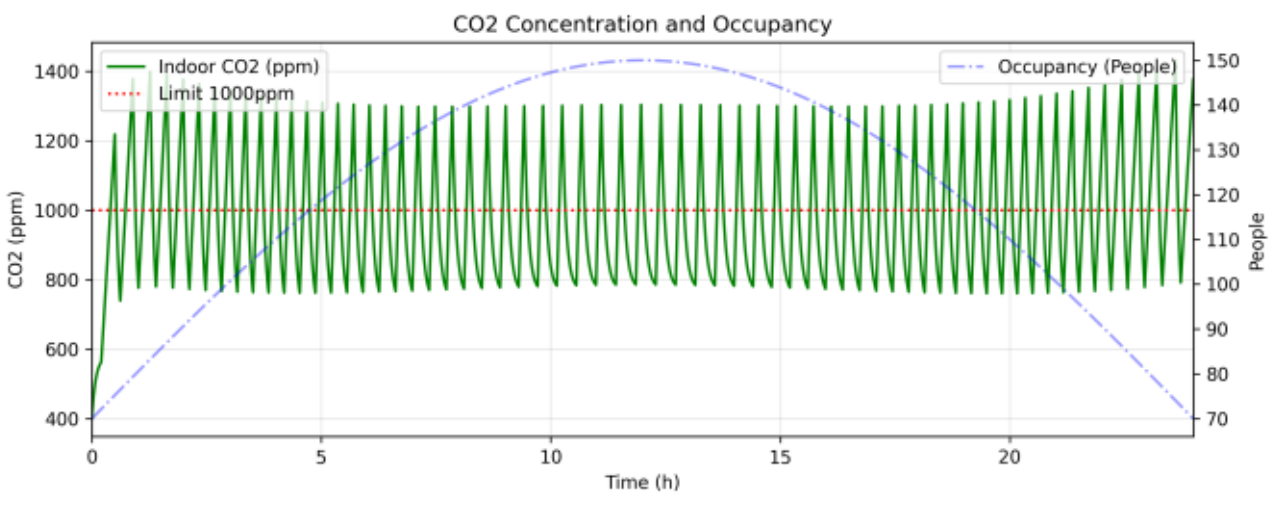

**a.** Thermostat controller

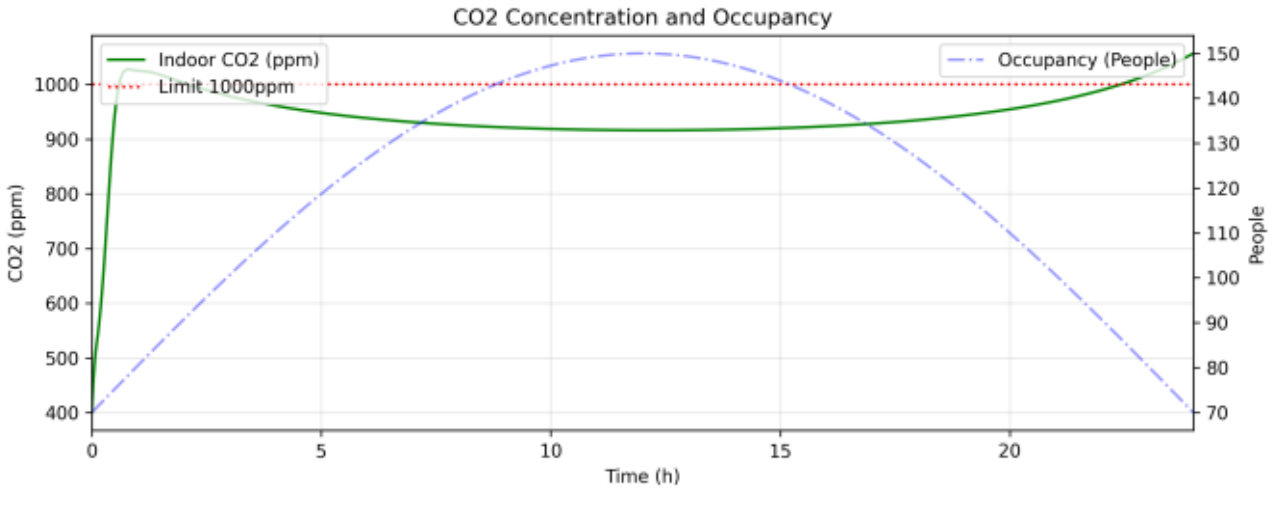

**b.** PID controller

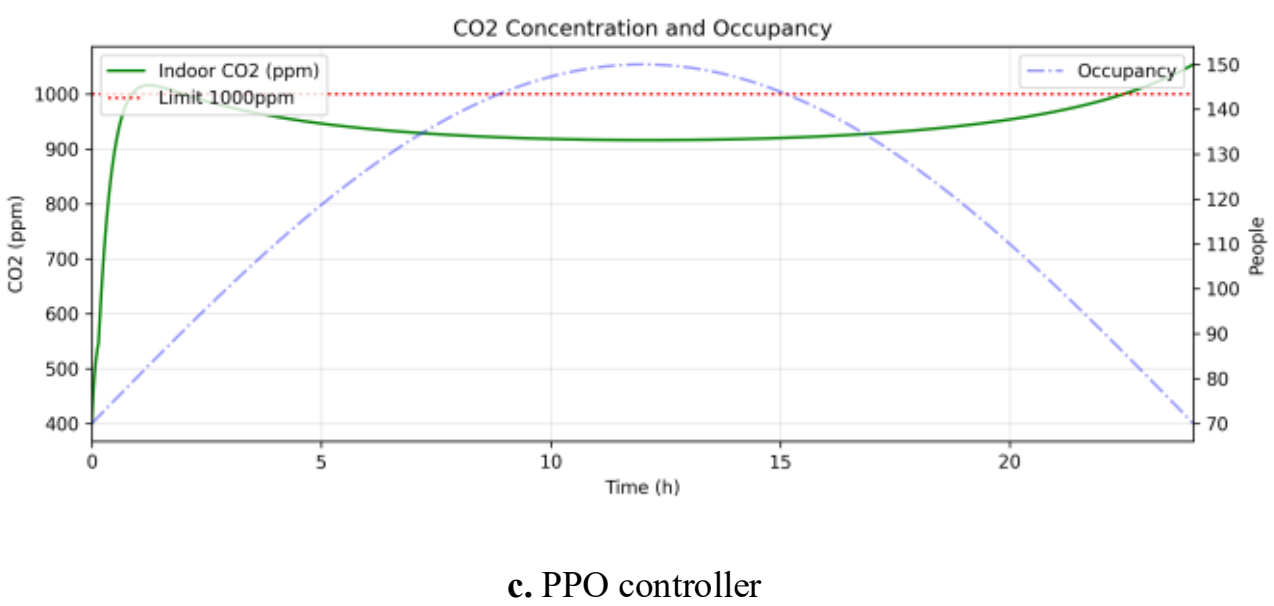

**c.** PPO controller

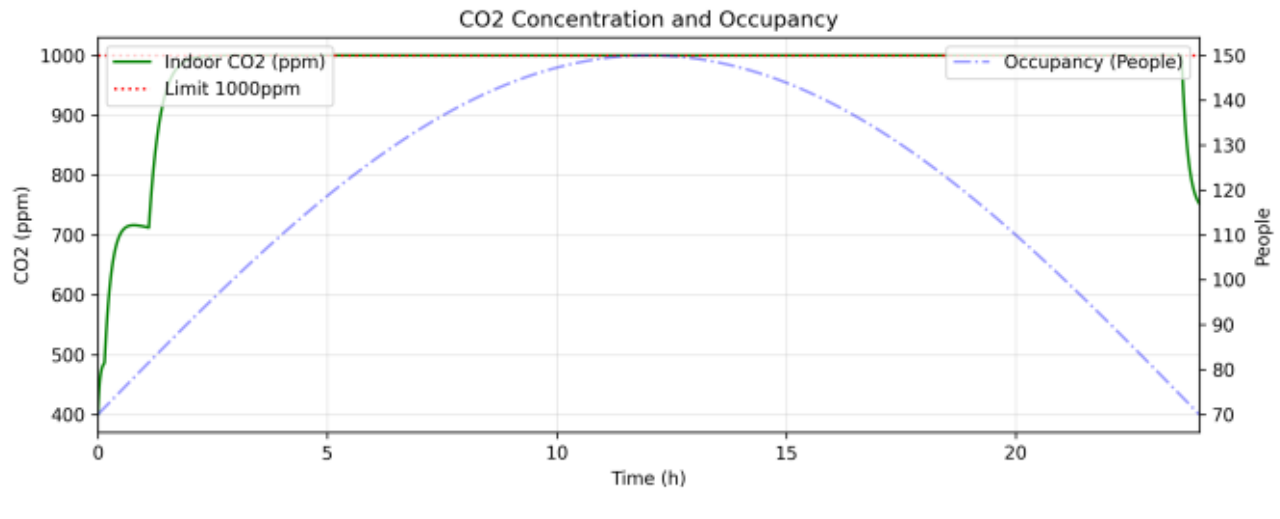

**d.** PPO controller with economizer

**Fig. 6.** Comparison of $CO_2$ Variations Under Different Controllers in space

6.3. Energy Consumption (Coil Power): Total Energy for Cooling consumption ($E_{total}$) is throughout the 24-hour period of simulation. The total cooling energy used for the On/Off Thermostat was (2776 kWh); for the PID controller it was (2760 kWh); for the average PPO, it was (2762 kWh); and for the smart PPO, it was (2609 kWh). In comparison with the On/Off and/or PID Controllers, the Smart Fresh Air Logic (Smart PPO) savings on energy are approximately 6%. In addition, these savings were achieved while still achieving best possible indoor air quality (i.e., lowest amount of $CO_2$). Furthermore, the Reinforcement Learning (RL) agent learned how much outside air to supply to meet health criteria, thereby reducing cooling load due to external air according to each appropriate amount necessary to achieve those calculations

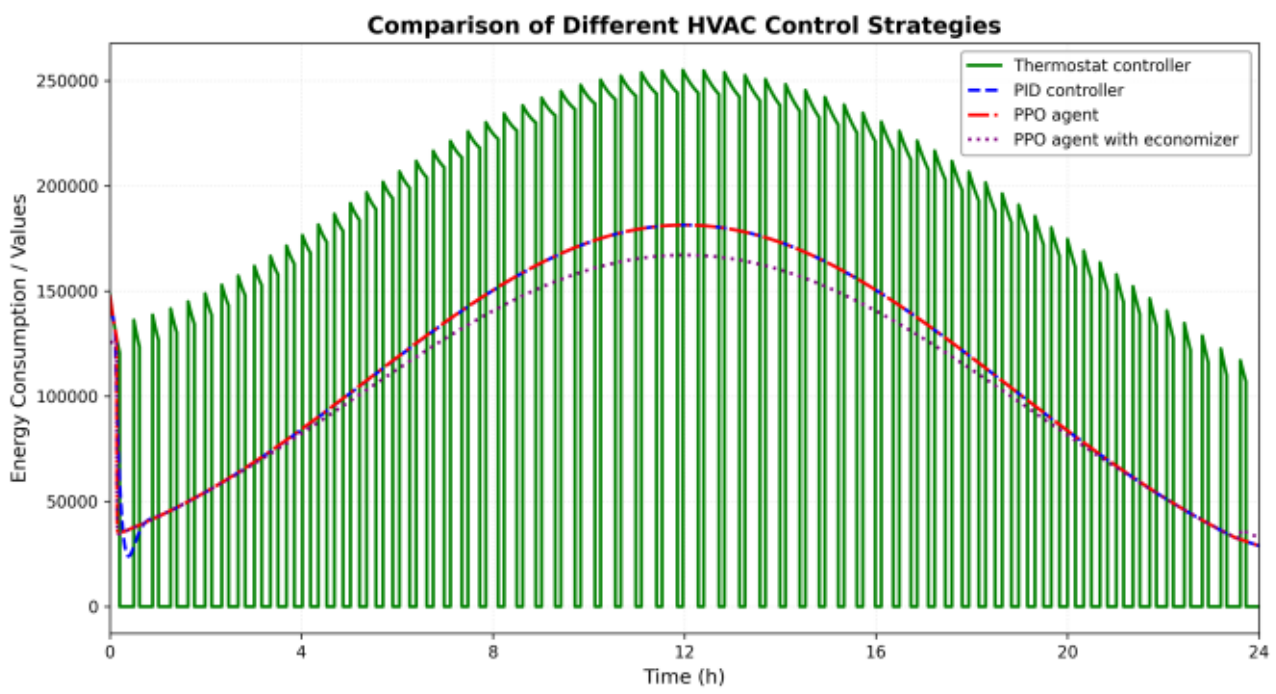

**Fig. 7.** Comparison of energy consumption in different control strategy

## VII. FUTURE WORK

In our future work, our ongoing work will be centered on advancing the previously developed modeling framework through two phases based on actual building data. The first phase will incorporate actual building data to develop a high-fidelity (i.e., precise) dynamic model in the MATLAB, Simulink, and Simscape environments. This dynamic model will represent the true thermal dynamics of the system so that we can use it as a basis for evaluating and validating the control performance of the PPO agent in real-world conditions. The second phase will involve deploying the reinforcement learning control strategy to a real, physical air handling unit (AHU) serving an actual building in order to demonstrate how the PPO agent can adapt to real-time conditions, provide energy-efficient performance, and operate reliably in live HVAC systems.

## Reference

[1] S. L. Zhou, A. A. Shah, P. K. Leung, X. Zhu, and Q. Liao, "A comprehensive review of the applications of machine learning for HVAC," *DeCarbon*, vol. 2, p. 100023, 2023, doi: 10.1016/j.decarbon.2023.100023.

[2] M. González-Torres, L. Pérez-Lombard, J. F. Coronel, I. R. Maestre, and D. Yan, "A review on buildings energy information: Trends, end-uses, fuels and drivers," *Energy Reports*, vol. 8, pp. 626–637, 2022, doi: 10.1016/j.egyr.2021.11.280.

[3] Md. W. Akram, M. F. Mohd Zublie, Md. Hasanuzzaman, and N. Abd Rahim, "Global Prospects, Advance Technologies and Policies of Energy-Saving and Sustainable Building Systems: A Review," *Sustainability*, vol. 14, no. 3, p. 1316, 2022, doi: 10.3390/su14031316.

[4] W. Shang, J. Liu, H. Meng, L. Jia, and X. Dai, "A RL-based human behavior oriented optimal ventilation strategy for better energy efficiency and indoor air quality," *Energy and Buildings*, vol. 345, p. 116072, 2025.

[5] Y. Boutahri and A. Tilioua, "Machine learning-based predictive model for thermal comfort and energy optimization in smart buildings," *Results in Engineering*, vol. 22, p. 102148, 2024, doi: 10.1016/j.rineng.2024.102148.

[6] J. Y. Yun *et al.*, "AI-Driven Non-Intrusive Occupant Monitoring for Occupant-Centric Control: A Multi-Domain Review of Thermal, Air, and Visual Environments," *Building and Environment*, 2026.

[7] F. Islam, I. Ahmed, and L. Mihet-Popa, "Development and testing of an IoT platform with smart algorithms for building energy management systems," *Energy and Buildings*, vol. 344, p. 115970, 2025.

[8] A. Chatterjee and D. Khovalyg, "Dynamic indoor thermal environment using Reinforcement Learning-based controls: Opportunities and challenges," *Building and Environment*, vol. 244, p. 110766, 2023.

[9] A. Chatterjee and D. Khovalyg, "Dynamic indoor thermal environment control using Reinforcement Learning: Balancing energy efficiency and human well-being," *Engineering Applications of Artificial Intelligence*, vol. 167, p. 113846, 2026, doi: 10.1016/j.engappai.2026.113846.

[10] S. Haghighat and R. Zhang, "Deep learning and digital twin integration for indoor environmental conditions (IEC): A state-of-the-art review," *Indoor Environments*, vol. 2, no. 4, p. 100137, 2025.

[11] J. Y. Yun *et al.*, "AI-Driven Non-Intrusive Occupant Monitoring for Occupant-Centric Control: A Multi-Domain Review of Thermal, Air, and Visual Environments," *Building and Environment*, vol. 299, p. 114673, 2026.

[12] M. Mohsenpour and Y. Xing, "Hybrid Reinforcement Learning for occupant-centric building control: A review and deployment framework for co-optimizing energy, comfort, and indoor air quality," *Applied Energy*, vol. 408, p. 127392, 2026.

[13] P. K. Hashir, S. Veerasingam, R. M. Haris, F. Sadooni, and S. Ghani, "A systematic review of artificial intelligence applications for indoor air quality monitoring in educational settings," *Engineering*

*Applications of Artificial Intelligence*, vol. 165, p. 113383, 2026, doi: 10.1016/j.engappai.2025.113383.
[14] L. B. Ahrendsen *et al.*, “Monte Carlo simulations and sensitivity analysis for developing control strategies in hybrid ventilation systems,” *Energy and Buildings*, vol. 359, p. 117294, 2026, doi: 10.1016/j.enbuild.2026.117294.
[15] H. Li, W. O’Brien, V. Loftness, E. C. Hameen, and T. Hong, “A critical review of use cases and insights from a large dataset of smart thermostats,” *Advances in Applied Energy*, vol. 19, p. 100236, 2025, doi: 10.1016/j.adapen.2025.100236.
[16] K. J. Åström and T. Hägglund, *Advanced PID Control*. ISA—The Instrumentation, Systems, and Automation Society, 2006.
[17] A. Afram and F. Janabi-Sharifi, “Theory and applications of HVAC control systems – A review of model predictive control (MPC),” *Building and Environment*, vol. 72, pp. 343–355, 2014.
[18] L. Bai and Z. Tan, “Optimizing energy efficiency, thermal comfort, and indoor air quality in HVAC systems using a robust DRL algorithm,” *Journal of Building Engineering*, vol. 98, p. 111493, 2024.